\documentclass[letterpaper]{article} %
\pdfoutput=1
\usepackage{aaai2026}  %
\usepackage{times}  %
\usepackage{helvet}  %
\usepackage{courier}  %
\usepackage[hyphens]{url}  %
\usepackage{graphicx} %
\usepackage{natbib}  %
\usepackage{caption} %
\usepackage{algorithm}
\usepackage{algorithmic}

\usepackage{newfloat}
\usepackage{listings}
\DeclareCaptionStyle{ruled}{labelfont=normalfont,labelsep=colon,strut=off} %
\floatstyle{ruled}
\newfloat{listing}{tb}{lst}{}
\floatname{listing}{Listing}
\usepackage{amssymb}  
\usepackage{amsmath} 
\usepackage{bm} 
\usepackage{diagbox}
\usepackage{tabularx}
\usepackage{multirow,makecell}
\usepackage[table]{xcolor} %
\usepackage{subfig}   %

\usepackage{nicematrix,booktabs}

\usepackage{cleveref}

\title{ConSurv: Multimodal Continual Learning for Survival Analysis}
\author{
    Dianzhi Yu\textsuperscript{\rm 1}, 
    Conghao Xiong\textsuperscript{\rm 1}, 
    Yankai Chen\textsuperscript{\rm 2}, 
    Wenqian Cui\textsuperscript{\rm 1}, 
    Xinni Zhang\textsuperscript{\rm 1}, 
    Yifei Zhang\textsuperscript{\rm 3}, \\
    Hao Chen\textsuperscript{\rm 4}, 
    Joseph J.Y. Sung\textsuperscript{\rm 3}, 
    Irwin King\textsuperscript{\rm 1} 
}
\affiliations{
    \textsuperscript{\rm 1}The Chinese University of Hong Kong\\
    \textsuperscript{\rm 2}University of Illinois Chicago\\
    \textsuperscript{\rm 3}Nanyang Technological University\\
    \textsuperscript{\rm 4}The Hong Kong University of Science and Technology\\

    \{dzyu23, chxiong21, wqcui23, xnzhang23, king\}@cse.cuhk.edu.hk,
    ychen588@uic.edu,\\
    \{yifei.zhang, josephsung\}@ntu.edu.sg,
    jhc@cse.ust.hk 
}

\usepackage{bibentry}

\begin{document}

\maketitle

\begin{abstract}
Survival prediction of cancers is crucial for clinical practice, as it informs mortality risks and influences treatment plans. 
However, a \textit{static} model trained on a single dataset fails to adapt to the \textit{dynamically evolving} clinical environment and continuous data streams, limiting its practical utility. 
While continual learning (CL) offers a solution to learn dynamically from new datasets, existing CL methods primarily focus on unimodal inputs and suffer from severe catastrophic forgetting in survival prediction. 
In real-world scenarios, multimodal inputs often provide comprehensive and complementary information, such as whole slide images and genomics; and neglecting inter-modal correlations negatively impacts the performance.
To address the two challenges of \textit{catastrophic forgetting} and \textit{complex inter-modal interactions} between gigapixel whole slide images and genomics, we propose \textbf{ConSurv}, the \textbf{first} multimodal continual learning (MMCL) method for survival analysis. ConSurv incorporates two key components: Multi-staged Mixture of Experts (MS-MoE) and Feature Constrained Replay (FCR). 
MS-MoE captures both task-shared and task-specific knowledge at different learning stages of the network, including two modality encoders and the modality fusion component, learning inter-modal relationships. 
FCR further enhances learned knowledge and mitigates forgetting by restricting feature deviation of previous data at different levels, including encoder-level features of two modalities and the fusion-level representations. 
Additionally, we introduce a new benchmark integrating four datasets, Multimodal Survival Analysis Incremental Learning (MSAIL), for comprehensive evaluation in the CL setting.
Extensive experiments demonstrate that ConSurv outperforms competing methods across multiple metrics. 
\end{abstract}

\begin{links}
    \link{Code}{https://github.com/LucyDYu/ConSurv}
\end{links}

\section{Introduction}
\label{sec: Introduction}

Survival prediction plays an important role in clinical practice, as it quantifies mortality risks and informs therapeutic decision-making.
Recent advances in deep learning have empowered researchers to make substantial progress in developing effective survival prediction models~\cite{Zadeh2020Bias}. 
Such efforts originally started with unimodal data sources like Whole Slide Images (WSIs)~\cite{Ilse2018Attentionbased, Lu2021Dataefficient, Shao2021Transmil, Zhang2022Dtfdmil} or genomics~\cite{Klambauer2017Selfnormalizing, Vaswani2017Attention, Chaudhary2018Deep}.
More recently, researchers have incorporated multimodal inputs, including both WSIs and genomics~\cite{Chen2021Multimodal, Chen2022Pancancer, Li2022HFBSurv, Zhou2023Crossmodal, Xu2023Multimodalb, Xiong2024MoME}, as they provide comprehensive and complementary information. For comprehensiveness, genomic features responsible for tumor formation correlate strongly with image patches of WSIs containing tumor cells~\cite{Chen2021Multimodal}. 
Regarding complementarity, WSIs are particularly valuable in late-stage cancer diagnosis, where survival outcomes are more predictable with morphological patterns.
Meanwhile, genomic data provides critical insights in early-stage cancer, where genetic factors drive tumor progression.

\begin{figure}[!t]
    \centering
    \includegraphics[width=8.5cm,page=1]{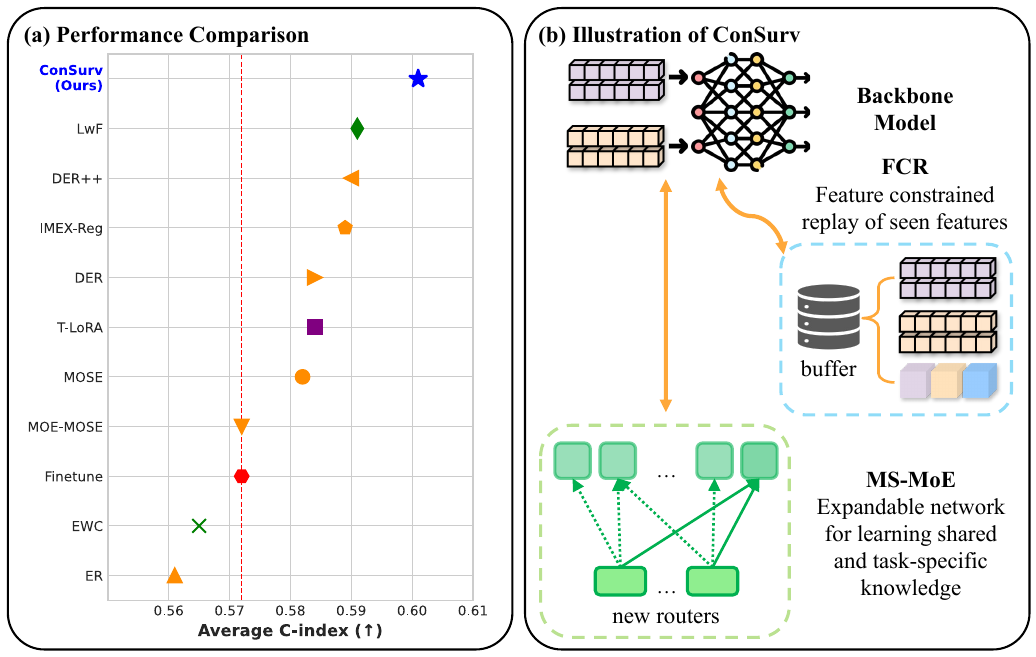}
    \caption{(a) Comparison of ConSurv against various CL methods on the MSAIL benchmark. (b) Illustration of ConSurv. It learns task-shared and task-specific multimodal knowledge during training with expandable MS-MoE, while consolidating previously acquired knowledge through FCR.
    A detailed architecture of ConSurv is presented in~\Cref{fig: Architecture}.}
    \label{fig: Inspiration for ConSurv}
\end{figure} 

However, given the \textit{dynamically changing} clinical environment, the \textit{static learning} paradigm, where a model is trained on a single data source, struggles to adapt and generalize to new data~\cite{Pianykh2020Continuous}. 
Such static models have limited sustainability and become outdated due to ongoing medical data collection, variations in staining protocols, and technological advancements that enhance the quality of WSI imaging and genomic data~\cite{Perkonigg2021Dynamic, Shen2022Federated, Huang2023ConSlide}. 
Moreover, data from different cancers often exhibit similar patterns~\cite{Baba2010Comparative}, and leveraging them appropriately enhances model performance~\cite{Xiong2024TAKT}. Consequently, it is both necessary and beneficial for models to learn from multiple datasets.
Although retraining a model on new and existing datasets together is a possible solution, it incurs significant computational time and high resource cost.

Continual learning (CL) offers a viable framework to overcome the limitations of static models and repeated retraining~\cite{Huang2023ConSlide, Yu2024Recent}, thus enhancing the practical utility and effectiveness of survival prediction models.
In the CL setting, a model is trained sequentially on multiple datasets, allowing it to adapt to new information without requiring explicit retraining on prior datasets.
A direct approach is finetuning the well-trained model on new datasets, but this strategy leads to \emph{catastrophic forgetting}, which means the model will suffer from severe performance decline on previously learned datasets~\cite{McCloskey1989Catastrophic, Ratcliff1990Connectionist}. This phenomenon occurs because parameters are updated to accommodate new knowledge and thereby deviate from the old optimal state for the previous datasets~\cite{Hassabis2017NeuroscienceInspired}. 
CL methods design mechanisms to mitigate catastrophic forgetting throughout the learning process.
CL aims to effectively balance \textit{plasticity}, the ability to acquire new knowledge continuously, and \textit{stability}, the capacity to retain previously learned information. This is referred to as the \textit{stability-plasticity dilemma}~\cite{Mermillod2013Stabilityplasticity, Masana2022ClassIncremental, Yu2024Recent}, where prioritizing the retention of previously learned knowledge can compromise the acquisition of new knowledge, as this process inevitably diminishes the specific knowledge essential for previous datasets.

While extensive research has explored CL for unimodal data~\cite{Li2017Learning,Kirkpatrick2017Overcoming,Wang2023Orthogonal,Huang2023ConSlide,Zhang2024Influential,WangModel}, work on multimodal data is limited~\cite{Yu2024Recent}. Specifically, no CL research focuses on WSIs and genomics modalities.
We find that directly applying unimodal CL methods to multimodal continual learning (MMCL) for survival prediction can lead to suboptimal performance. 
We identify two challenges: 
(C1) \emph{Severe catastrophic forgetting}.
As shown in Figure~1a, simple adaptations in this setting are not ideal, and sometimes even yield worse results than vanilla sequential finetuning, highlighting the severe catastrophic forgetting that current CL methods still face. 
(C2) \emph{Complex inter-modal interactions}.
While WSIs and genomic data offer complementary information, the model needs to effectively learn the complex correlations between these modalities~\cite{Xu2023Multimodal}, especially when these correlations vary across different datasets, which previous CL methods have neglected. 
Notably, these two challenges are intertwined, influencing and exacerbating each other. 
Catastrophic forgetting can be more severe when distinct modalities of WSIs and genomics are involved, due to different data sizes, inconsistent distributions and representations that arise from data heterogeneity~\cite{baltrusaitis2017multimodalmachinelearningsurvey, Peng2021Hierarchical, Yu2024Recent}.
The learned multimodal interactions diminish due to catastrophic forgetting when the model concentrates on learning new cancer datasets.

To tackle the challenges above, we propose \textbf{Con}tinual \textbf{Surv}ival Analysis (\textbf{ConSurv}), a novel MMCL method designed to learn complex correlations from WSIs and genomics data and preserve previously acquired information throughout the CL process. 
We propose an expandable Multi-staged Mixture of Experts (MS-MoE) module (see~Figure~1b).
It facilitates the modeling of \textit{complex inter-modal relationships} during continual training by flexibly combining different experts. 
It captures both shared and task-specific knowledge from datasets at different multimodal learning stages within the model, specifically WSI and genomic encoders, and the fusion component. As new datasets are introduced, corresponding new routers are added to select relevant experts, which aids in mitigating catastrophic forgetting.

To further \textit{enhance the retention of previous knowledge and mitigate forgetting}, we introduce Feature Constrained Replay (FCR). Since directly storing large WSIs and genomic data induces large storage costs, we store only the processed instance feature representations for replay. 
During training, FCR constrains the deviation of individual features of both modalities after their respective encoders, and the final fused representations of previous datasets to alleviate forgetting through knowledge distillation~\cite{Gou2021Knowledge}.

Acknowledging the absence of benchmarks in MMCL for survival prediction using WSIs and genomics, we propose a new challenging \textbf{M}ultimodal \textbf{S}urvival \textbf{A}nalysis \textbf{I}ncremental \textbf{L}earning (\textbf{MSAIL}) benchmark to evaluate various CL methods. 
This benchmark utilizes four publicly available survival prediction datasets from The Cancer Genome Atlas Program (TCGA\footnote{\url{https://www.cancer.gov/ccg/research/genome-sequencing/tcga}}), namely BLCA, UCEC, LUAD, and BRCA.
We evaluate our ConSurv on the MSAIL benchmark, and it outperforms other methods on multiple metrics through extensive experiments. 
We emphasize that ConSurv successfully achieves an effective balance in the stability-plasticity trade-off, effectively acquiring new knowledge while retaining previously learned information.

\begin{figure*}[!t]
    \centering
    \includegraphics[width=17.7cm,page=2]{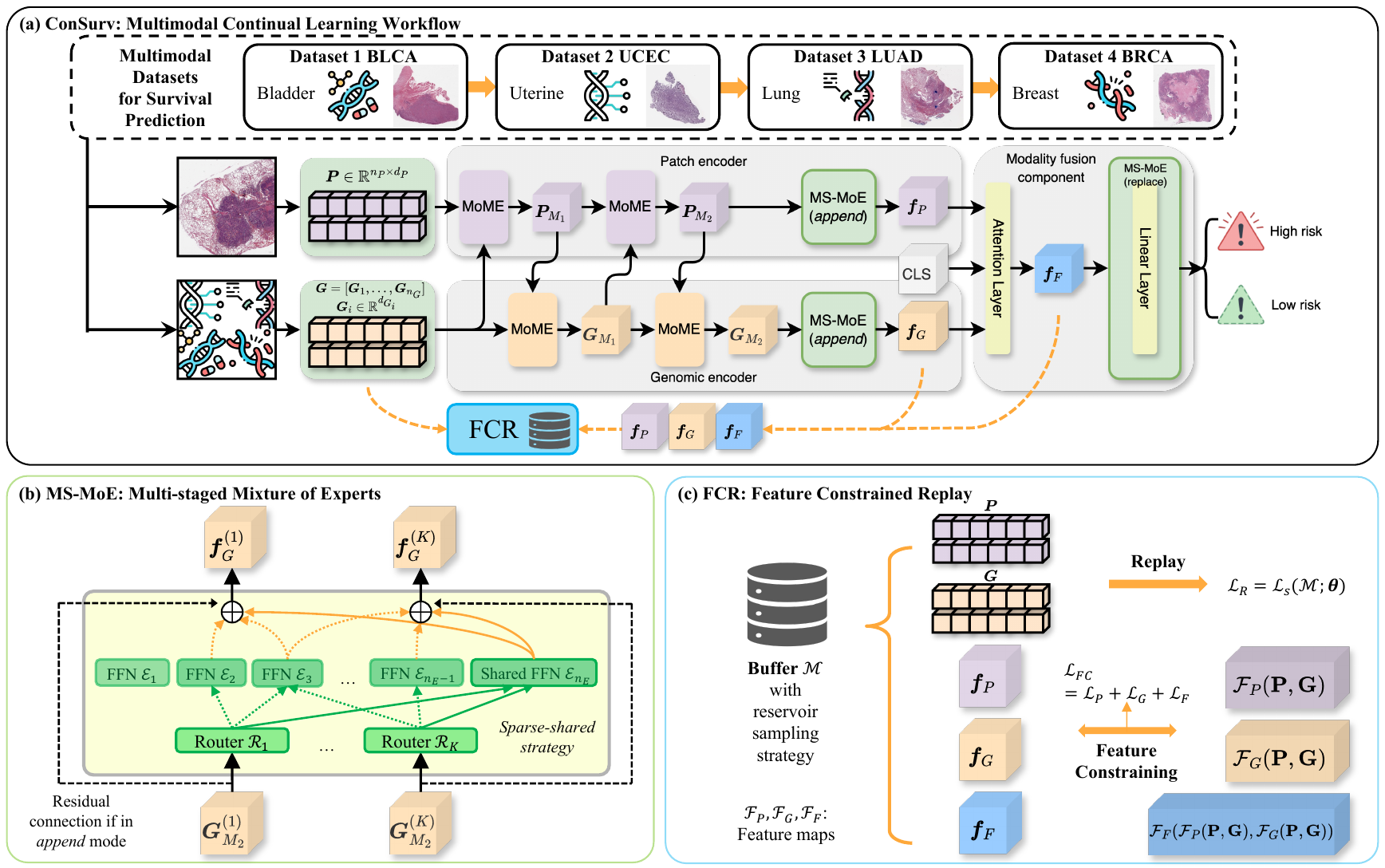}
    \caption{Overall architecture of ConSurv. 
    (a) The MMCL workflow for continual survival prediction across different cancer datasets. We employ a recent SOTA model, MoME~\cite{Xiong2024MoME}, in survival prediction as our backbone model. We train the model sequentially on the multimodal datasets. 
    (b) MS-MoE learns both shared and task-specific knowledge at different learning stages of the network, including WSI and genomic encoders and the modality fusion component.
    (c) FCR preserves previously learned knowledge through additional loss terms on the replay buffer. }
    \label{fig: Architecture}
\end{figure*}

The contributions of our paper are summarized as follows:
\begin{enumerate}
	\item We propose ConSurv, the \textbf{first} MMCL method for survival analysis across multiple cancers using WSI and genomic data.
  	\item We design MS-MoE to handle complex, dynamic inter-modal interactions, and FCR to alleviate catastrophic forgetting.
    \item We propose a new MSAIL benchmark for evaluation, covering datasets of four cancers from TCGA. 
    \item The extensive experiments on MSAIL demonstrate not only the superiority of ConSurv over competing methods, but also the effectiveness of the proposed modules.
\end{enumerate}

\section{ConSurv Methodology}

\subsection{MMCL Workflow}
MMCL for survival prediction is a \textit{CL} setting where a model sequentially learns to predict the survival time on \textit{multimodal} datasets. We provide detailed discussions on related work and preliminaries for survival prediction in Appendix A and B, respectively.
We define a \textit{multimodal dataset sequence} $\mathcal{D} = \{\mathcal{D}_1, \mathcal{D}_2, \ldots, \mathcal{D}_K\}$, where $\mathcal{D}_i$ are different multimodal cancer datasets and $K$ is the number of datasets.
In the MMCL setting, the model is trained on the current dataset $\mathcal{D}_k$ (optionally with limited access to previous datasets).
During training, the model parameters $\bm{\theta}$ are updated in a controlled manner, facilitating learning on $\mathcal{D}_k$ while mitigating the forgetting of knowledge learned from $\{\mathcal{D}_1, \mathcal{D}_2, \cdots, \mathcal{D}_{k-1}\}$.
The CL loss of dataset $k$ has the following general form~\cite{WangModel}:
\begin{equation}
    \mathcal{L}_{CL}^{(k)}(\bm{\theta})=\mathcal{L}_s(\mathcal{D}_k;\bm{\theta})+\zeta \mathcal{L}_f^{(k)}(\bm{\theta}),
\end{equation}
where $\mathcal{L}_s$ is the loss for survival prediction, applying to the current dataset $\mathcal{D}_k$; 
$\mathcal{L}_f$ is a forgetting-mitigation term, such as memory-replay and parameter-regularization loss;
$\zeta$ denotes a constant that balances knowledge acquisition and forgetting avoidance in the stability-plasticity trade-off~\cite{WangModel}.
Model minimizes $\mathcal{L}_{CL}^{(k)}$ when training on each respective dataset $\mathcal{D}_k$, i.e., 
\mbox{$\bm{\theta}^{*(k)}=\operatorname{argmin}_{\bm{\theta}} \ \mathcal{L}_{CL}^{(k)}(\bm{\theta}).$}
In terms of model performance, CL aims to 
obtain a model which achieves high performance on all trained datasets, i.e., 
\mbox{$\bm{\theta}^{*}=\operatorname{argmax}_{\bm{\theta}}  \sum_{i=1}^{K} P(\bm{\theta}, \mathcal{D}_i),$}
where $P(\bm{\theta}, \mathcal{D}_i)$ is the performance evaluation function based on different types of datasets and tasks. In the case of survival prediction, the function is typically the Concordance index (C-index) and C-index IPCW (inverse probability of censoring weights)~\cite{Uno2011Statistics}. 

We focus on the task-incremental learning (TIL) setting in this work~\cite{vandeVen2022Three}. In the CL literature, the term ``task'' directly corresponds to its dataset, and therefore, terms ``task'' and ``dataset'' are often used interchangeably. 
In TIL, for any two distinct datasets $i \neq j$, $\mathcal{D}_i$ and $\mathcal{D}_j$ exhibit different input distributions and label spaces~\cite{Wang2024Comprehensive}.
This aligns with our setting, where multimodal datasets correspond to different cancer types and exhibit different survival time distributions.
Although we partition the survival time of each dataset into the same number of bins, the time intervals of bins are all different, hence different label spaces.
Task identities are available at inference. 
We employ MoME~\cite{Xiong2024MoME} as the multimodal backbone and apply our proposed modules MS-MoE and FCR in the MMCL training workflow (Sections~\ref{sec: Methodology MS-MoE} and \ref{sec: FCR}). 
We choose MoME because it is a recent SOTA model in survival prediction, whose architectural design aligns with the concept of MoE, using routers to select different experts to fuse multimodal inputs.

\subsection{MS-MoE: Multi-staged Mixture of Experts}
\label{sec: Methodology MS-MoE}

To address the challenge of learning dynamic inter-modal interactions between WSIs and genomics data in cancer datasets, we introduce Multi-staged Mixture of Experts (MS-MoE). It is an expandable module designed for integration with the multimodal backbone while preserving the backbone's core architecture (Figure~2a and 2b). 
We adopt and modify the Mixture of Experts (MoE)~\cite{Shazeer2017Outrageously} as the base component of MS-MoE. 
\paragraph{Routing Mechanism.}
The original MoE comprises a set of ``expert'' subnetworks and a ``router'' that selects them.
Instead of adding new experts for each new cancer dataset $\mathcal{D}_k$, we keep a fixed number $n_E$ of experts $\{\mathcal{E}_i\}_{i=1}^{n_E}$ and only introduce a new linear-layer router $\mathcal{R}_k$, to limit parameter growth, following~\cite{Yu2024Boosting}. 
Using task-specific routers helps to \textit{mitigate catastrophic forgetting} when learning a new cancer type.
We employ Sparse MoE~\cite{Jiang2024Mixtrala, Fedus2022Review} to selectively utilize experts instead of always using all experts.
This strategy reduces computational costs and encourages experts to learn task-specific knowledge.
Additionally, when the same expert is selected for inputs from a subset of tasks, this strategy facilitates inter-task collaboration and the learning of shared knowledge.
We visualize and support the above claims in~\Cref{sec: Routing Analysis}.
To enable the module to acquire knowledge across all datasets, we designate one expert as a shared expert, ensuring it remains consistently active, inspired by~\cite{pmlr-v162-rajbhandari22a}.
The gating weights $W^{(k)}$ employing this \textit{sparse-shared strategy} are defined as:
\begin{equation}
    W^{(k)}=\text{Softmax}\left(\text{TopK-S}\left(\mathcal{R}_k(\mathbf{x}^{(k)})\right)\right),
\end{equation}
where $\text{TopK-S}(\cdot)$ selects the shared expert and top $k$ experts among the rest experts, while setting non-selected ones to be $-\infty$. $\text{Softmax}(\cdot)$ normalizes the weights. For input $\mathbf{x}^{(k)}$ when training on $\mathcal{D}_k$, the output $\mathbf{y}^{(k)}$ of the MoE module $\mathcal{MS}$ is defined as:
\begin{equation}
    \mathbf{y}^{(k)}=\mathcal{MS}(\mathbf{x}^{(k)})=\sum_{i=1}^{n_E} W^{(k)}_i \mathcal{E}_i(\mathbf{x}^{(k)}),
\end{equation}
where $W^{(k)}_i$ denotes the $i$-th entry of $W^{(k)}$.

\paragraph{Integration into Backbone.}
Consistent with our goal of integrating the above MS-MoE modules into the backbone model while maintaining its core structure, we introduce two integration modes: \textit{replace} and \textit{append}.
If the target insertion point within the MoME architecture contains a standard Feed-Forward Network (FFN) or linear layer, we \textit{replace} this component entirely with an MS-MoE module.
The experts $\{\mathcal{E}_i\}$ within this module are configured to have an architecture identical to that of the replaced component. 
In this mode, if the learned gating weights strongly favor the shared expert (i.e., weight close to $1$), the computation approximates that of the original layer, and the modified structure effectively reduces to the original backbone.
If the target insertion point lacks such layers to replace, we \textit{append} the MS-MoE module residually. 
We employ two-layer FFNs as the experts $\{\mathcal{E}_i\}$.
The output $\mathbf{y}^{(k)}$ incorporates a residual connection~\cite{He2016Deep} from the input: $\mathbf{y}^{(k)}=\mathbf{x}^{(k)}+\mathcal{MS}(\mathbf{x}^{(k)}).$
In this configuration, if the activated experts learn negligible transformations (effectively mapping inputs close to zero), the output $\mathbf{y}^{(k)}$ approximates the original input $\mathbf{x}^{(k)}$, again preserving the backbone's information flow.
Consequently, these integration strategies allow MS-MoE to introduce additional capacity and flexibility for CL while maintaining the integrity and performance baseline of the original MoME architecture.

\paragraph{Capturing Inter-modal Interactions.}
To effectively capture and adapt the \textit{complex inter-modal interactions} between WSI and genomic data, MS-MoE modules are strategically placed at the end of the WSI encoder, the genomic encoder, and the modality fusion component of the MoME backbone.
As noted in~\cite{Xiong2024MoME}, MoME's encoders already perform cross-modal processing (e.g., the patch encoder uses genomic information, and vice-versa). 
By adding MS-MoE at all three key stages, our approach aims to enhance the learning of these multimodal interactions while maintaining CL capabilities, adapting expert knowledge and task-specific routing as new cancer datasets are encountered.

\begin{table*}[!ht]
\centering

\begin{NiceTabular}{w{c}{1cm}|llllll}
\toprule
\diagbox{Train}{Val}                  & BLCA      & UCEC      & LUAD      & BRCA      & Average (on trained) & Random \\
\midrule
BLCA & \textbf{0.607±0.026} &         0.545±0.025            & -                    & -                    & 0.607±0.026    &  0.519±0.052 \\
UCEC & 0.500±0.055          & \textbf{0.648±0.080} &  0.481±0.045                 & -                    & 0.574±0.061    &  0.420±0.024 \\
LUAD & 0.512±0.046          & 0.607±0.076          & \textbf{0.642±0.046} & 0.518±0.044                  & 0.587±0.036     & 0.475±0.073 \\
BRCA & 0.531±0.035          & 0.588±0.062          & 0.519±0.041          & \textbf{0.650±0.059} & 0.572±0.024  & 0.473±0.089 \\
\bottomrule

\end{NiceTabular}
\caption{Performance (C-index) of each dataset under sequential finetuning.}
\label{tab: RQ1}
\end{table*}

\subsection{FCR: Feature Constrained Replay}
\label{sec: FCR}

To further \textit{alleviate catastrophic forgetting}, we introduce Feature Constrained Replay (FCR), designed to retain previously acquired knowledge when training on multimodal datasets, as depicted in Figure~2c. 
The FCR module maintains features of a small fixed number of seen instances. During training, it constrains the deviation of features from WSIs and genomics data after their respective encoders and the final fused representations of previous datasets to mitigate forgetting.
Let $\mathcal{F}_P$, $\mathcal{F}_G$, and $\mathcal{F}_F$ denote the feature maps of the WSI patch encoder, genomic encoder, and the final fusion component, respectively. Given WSI patches and genomic data $(\mathbf{P}, \mathbf{G})$, we obtain patch feature representation $\mathbf{f}_P = \mathcal{F}_P(\mathbf{P}, \mathbf{G})$, genomic feature representation $\mathbf{f}_G = \mathcal{F}_G(\mathbf{P}, \mathbf{G})$, and the final fusion representation $\mathbf{f}_F = \mathcal{F}_F(\mathbf{f}_P, \mathbf{f}_G)$.

We introduce a fixed-size replay buffer $\mathcal{M}$ and utilize a reservoir sampling strategy~\cite{Vitter1985Random} to randomly select sample features from the input data stream and update the buffer, ensuring equal retention probability for all seen instances. The buffer maintains three types of features: $\mathbf{f}_P, \mathbf{f}_G$, and $\mathbf{f}_F$.
For the final fusion representations, an $\mathcal{L}_2$ loss is utilized as a feature distillation technique.
It minimizes the distance between the representation $\mathbf{f}_F$ from the buffer $\mathcal{M}$ and that from the current model by following~\cite{Gou2021Knowledge, Bai2023Revisiting}:
\begin{align}
\mathcal{L}_{F}(\mathcal{M};\bm{\theta}) = & \ \mathbb{E}_{(\mathbf{P}, \mathbf{G}, \mathbf{f}_F) \sim \mathcal{M}} [\lVert \mathbf{f}_F - \\ \notag
& \mathcal{F}_F(\mathcal{F}_P(\mathbf{P}, \mathbf{G}),\mathcal{F}_G(\mathbf{P}, \mathbf{G})) \rVert_2^2 ].
\end{align}
We define the loss for patch feature representations as:
\begin{align}
    \mathcal{L}_{P}(\mathcal{M};\bm{\theta}) =
\mathbb{E}_{(\mathbf{P}, \mathbf{G}, \mathbf{f}_P) \sim \mathcal{M}} [\lVert \mathbf{f}_P - \mathcal{F}_P(\mathbf{P}, \mathbf{G}) \rVert_2^2 ],
\end{align}
and similarly for $\mathcal{L}_{G}(\mathcal{M};\bm{\theta})$. The total feature constraint loss is then given by:
\begin{equation}
    \mathcal{L}_{FC}(\mathcal{M};\bm{\theta}) = \mathcal{L}_{P}(\mathcal{M};\bm{\theta}) + \mathcal{L}_{G}(\mathcal{M};\bm{\theta}) + \mathcal{L}_{F}(\mathcal{M};\bm{\theta}).
\end{equation}

Furthermore, to leverage the replay buffer, the model is trained on the data points within the buffer using their ground truth labels, following ER~\cite{Chaudhry2019Tiny} and DER++~\cite{Buzzega2020Dark}. This additional replay loss is denoted as $\mathcal{L}_{R}(\mathcal{M};\bm{\theta})=\mathcal{L}_s(\mathcal{M};\bm{\theta})$. Consequently, during training on dataset $\mathcal{D}_k$, the overall loss for the model incorporating FCR is: 
\begin{equation}
    \mathcal{L}_{CL}^{(k)}(\bm{\theta})=\mathcal{L}_s(\mathcal{D}_k;\bm{\theta}) + \alpha \mathcal{L}_{FC}(\mathcal{M};\bm{\theta}) + \beta \mathcal{L}_{R}(\mathcal{M};\bm{\theta}),
\end{equation}
where $\alpha$ and $\beta$ are hyperparameters that control the relative weights of the losses, enabling a balance between learning from the current dataset $\mathcal{D}_k$ and preserving previous knowledge. 
Note that here $\mathcal{L}_{FC}$ and $\mathcal{L}_{R}$ are not superscripted by $k$ because their values depend on the update of buffer $\mathcal{M}$ throughout the training process.

\setlength{\tabcolsep}{3pt}
\begin{table*}[!t]
    \centering
\small
\begin{NiceTabular}{c|l|c|ccc|c|ccc}
    \CodeBefore
    \rowcolors{14}{gray!20}{}[cols=2-10,restart]
    \Body
    \toprule
    \multicolumn{1}{c|}{\multirow{2.5}{*}{Type}} & \multicolumn{1}{c|}{\multirow{2.5}{*}{Method}} & \multicolumn{4}{c|}{C-index}                                                               & \multicolumn{4}{c}{C-index IPCW}                                                        \\
    \cline{3-10}\addlinespace[0.5ex]
                          &                   & \textbf{Average} ($\uparrow$)            & Forget ($\downarrow$)         & BWT ($\uparrow$)                & FWT ($\uparrow$)                   & \textbf{Average} ($\uparrow$)              & Forget ($\downarrow$)            & BWT ($\uparrow$)                   & FWT ($\uparrow$)                  \\
    \midrule
    \multirowcell{2}{Base-\\line}   & Joint                                       & 0.611±0.037          & -                    & -                     & -                    & 0.545±0.045          & -                    & -                    & -                    \\
  & Finetune                                  & 0.572±0.024          & 0.094±0.041          & -0.086±0.047          & 0.058±0.007          & 0.528±0.055          & 0.136±0.091          & -0.101±0.108         & 0.022±0.160          \\
  \midrule
  \multirowcell{2}{Reg.}   & EWC                                        & 0.565±0.055          & 0.115±0.068          & -0.111±0.072          & 0.050±0.021          & 0.510±0.067          & 0.105±0.093          & -0.084±0.117         & 0.007±0.130          \\
  & LwF                                        & 0.591±0.034          & 0.072±0.039 & -0.065±0.047 & 0.040±0.043          & 0.589±0.029          & 0.065±0.056          & -0.019±0.093         & 0.020±0.156          \\
  \midrule    
  \multirowcell{1}{Arch.}   & T-LoRA                                        & 0.584±0.022         & \textbf{0.000±0.004}          & \textbf{0.002±0.005}          & 0.048±0.047          & 0.553±0.046          & 0.053±0.042          & -0.022±0.052         & 0.056±0.141          \\
  \midrule    
  {\cellcolor{white} \multirowcell{7}{Re-\\play}}                                         & ER                                         & 0.561±0.025          & 0.110±0.052          & -0.110±0.052          & 0.014±0.031          & 0.511±0.057          & 0.107±0.056          & -0.088±0.051         & -0.046±0.111         \\
    & DER                                                                  & 0.584±0.017          & 0.100±0.041          & -0.098±0.043          & 0.074±0.042          & 0.544±0.024          & 0.107±0.031          & -0.077±0.082         & 0.037±0.165          \\
    & DER++                                       & 0.590±0.030          & 0.082±0.019          & -0.074±0.009          & \textbf{0.084±0.026} & 0.541±0.074          & 0.102±0.097          & -0.074±0.114         & 0.055±0.083          \\
    & MOSE & 0.582±0.028 & 0.090±0.024 & -0.090±0.024 & -0.008±0.071 & 0.548±0.065 & 0.091±0.061 & -0.074±0.041 & -0.027±0.133 \\
    & MOE-MOSE & 0.572±0.023 & 0.116±0.040 & -0.113±0.044 & -0.037±0.087 & 0.547±0.054 & 0.109±0.075 & -0.077±0.052 & -0.011±0.163 \\
    & IMEX-Reg                                       & 0.589±0.042          & 0.092±0.036          & -0.087±0.037          & 0.048±0.049 & 0.589±0.032          & \textbf{0.037±0.070}          & -0.009±0.068         & 0.054±0.139          \\
    & \textbf{ConSurv}                              & \textbf{0.601±0.045} & 0.088±0.052          & -0.081±0.060          & 0.067±0.059          & \textbf{0.597±0.039} & 0.049±0.048 & \textbf{0.002±0.080} & \textbf{0.083±0.103}\\
\bottomrule
                                                
    \end{NiceTabular}
    \caption{Comparison results among different CL methods. The best performances are highlighted in bold. 
The main metrics are average C-index and average C-index IPCW. Forgetting, BWT, and FWT are reported for reference.}
    \label{tab: Main result}
    \end{table*}

\setlength{\tabcolsep}{4pt}
\begin{table*}[t]
    \centering
\small
    \begin{NiceTabular}{c|c|c|ccc|c|ccc}
    \toprule

    \multicolumn{1}{c|}{\multirow{2.5}{*}{FCR}} &  \multirowcell{2.5}{MS-\\MoE} & \multicolumn{4}{c|}{C-index}                                                               & \multicolumn{4}{c}{C-index IPCW}                                                     \\ \cline{3-10}\addlinespace[0.5ex]   
               &                 & \textbf{Average} ($\uparrow$)            & Forget ($\downarrow$)         & BWT ($\uparrow$)                & FWT ($\uparrow$)                   & \textbf{Average} ($\uparrow$)              & Forget ($\downarrow$)            & BWT ($\uparrow$)                   & FWT ($\uparrow$)                \\

    \midrule

                                             &                                             & 0.572±0.024          & 0.094±0.041          & -0.086±0.047          & 0.058±0.007          & 0.528±0.055          & 0.136±0.091          & -0.101±0.108         & 0.022±0.160          \\
                                             \checkmark                                  &  & 0.581±0.043 & 0.121±0.061 & -0.119±0.061 & \textbf{0.079±0.059} & 0.545±0.045          & 0.121±0.096          & -0.112±0.104         & 0.042±0.138          \\
                                             & \checkmark                                          & 0.585±0.022          & \textbf{0.079±0.034} & \textbf{-0.079±0.034} & 0.041±0.022          & 0.575±0.015          & 0.078±0.059          & -0.065±0.054         & 0.067±0.087          \\
                                             \checkmark (f)                                       & \checkmark                        & 0.599±0.026          & 0.083±0.030 & -0.082±0.030          & 0.044±0.051          & 0.554±0.045          & 0.086±0.104          & -0.060±0.093         & -0.022±0.111         \\
                                             \rowcolor{gray!20} \checkmark                                      & \checkmark                                             & \textbf{0.601±0.045} & 0.088±0.052          & -0.081±0.060 & 0.067±0.059          & \textbf{0.597±0.039} & \textbf{0.049±0.048} & \textbf{0.002±0.080} & \textbf{0.083±0.103} \\
    \bottomrule                                         
    \end{NiceTabular}
    \caption{Ablation study of FCR and MS-MoE in ConSurv.  ``$\checkmark$(f)'' denotes FCR with only the final fusion representation, as opposed to all three feature levels.
    }
    \label{tab: Ablation table}
\end{table*}

\section{Experiments}
\label{sec: Experiment}

\subsection{MSAIL Benchmark}
\label{sec: Metrics}

We propose a new benchmark, namely \textbf{M}ultimodal \textbf{S}urvival \textbf{A}nalysis \textbf{I}ncremental \textbf{L}earning (MSAIL) to evaluate different CL methods. Experimental settings, evaluation protocol
and implementation details are provided in Appendix C.

\subsubsection{Data.} 
Our MSAIL benchmark consists of four multimodal survival analysis datasets, which we collectively refer to as \textbf{Cancer4} for brevity in the context of continual training. 
These datasets are from The Cancer Genome Atlas Program (TCGA). 
Specifically, they are
Bladder Urothelial CArcinoma (BLCA) ($n$ = 373),
Uterine Corpus Endometrial Carcinoma (UCEC) ($n$ = 480),
LUng ADenocarcinoma (LUAD) ($n$ = 453),
and BReast Invasive CArcinoma (BRCA) ($n$ = 955).
The task order used for this benchmark is BLCA, UCEC, LUAD, and BRCA. 
We present an alternative task order and the results in Appendix C.4.

\subsubsection{Metrics.}
To evaluate performance on individual datasets, we employ the Concordance index (C-index) as our evaluation metric.
We moreover utilize C-index IPCW (inverse probability of censoring weights)~\cite{Uno2011Statistics} as another metric, which adjusts the bias introduced by censoring.
To evaluate the model in the MMCL setting, we compute \textit{Average Performance} of the above two metrics as the main metrics. 
We additionally report \textit{Forgetting}~\cite{Chaudhry2018Riemannian}, \textit{Backward Transfer} (BWT) and \textit{Forward Transfer} (FWT)~\cite{Lopez-Paz2017Gradient} for reference.

\subsection{Research Questions}
We aim to answer the following research questions:
\begin{itemize}
    \item \textbf{RQ1}: Motivation for CL. 
    \textbf{(1)} Does CL offer a performance benefit over a static model on new datasets? 
    \textbf{(2)} What is the severity of catastrophic forgetting with direct sequential finetuning?
    \item \textbf{RQ2}: Performance comparison. How effective is our proposed ConSurv compared with other CL methods?
    \item \textbf{RQ3}: In-depth Analysis of ConSurv. 
    \textbf{(1)} How effectively does ConSurv stratify patients into risk groups on each cancer dataset?
    \textbf{(2)} How does each component of ConSurv impact the performance?
    \textbf{(3)} How effective is the routing mechanism of MS-MoE for expert selection?
\end{itemize}

\subsection{Experimental Results}
\label{sec: Experimental Results}
\subsubsection{Necessity of continual learning (RQ1.1).}
We first explore whether a static model can generalize to new datasets.
We directly evaluate the initial model on all tasks before training, as the random performance baseline.
As shown in~\Cref{tab: RQ1}, a model trained on BLCA achieves a C-index of 0.607. We then evaluate it on the next dataset, UCEC, before training, and the performance is 0.545, which exceeds the random performance. This positive forward knowledge transfer is consistently observed when evaluating on each subsequent task, indicating the acquisition of shared knowledge.
However, this transferred performance is significantly lower than the performance after training on the respective dataset (e.g., 0.648 for UCEC), thus highlighting the necessity for CL approaches to achieve optimal performance on new datasets. 
Importantly, the existence of positive forward knowledge transfer enhances the efficacy of CL by providing an informed starting point for subsequent training.

\subsubsection{Quantification of catastrophic forgetting (RQ1.2).}
A common baseline in CL is to finetuning the model on new datasets.
We next examine the presence of catastrophic forgetting under this new setting. 
As evidenced in~\Cref{tab: RQ1}, the performance on the initially learned task drops dramatically from 0.607 down to 0.531 when the BLCA-trained model is subsequently trained on UCEC, LUAD, and BRCA. 
This phenomenon of forgetting previously learned tasks is consistently observed throughout the training sequence, demonstrating severe catastrophic forgetting.

\subsubsection{Comparison with other CL methods (RQ2).}
We compare ConSurv with finetuning and other SOTA unimodal CL methods in~\Cref{tab: Main result}.
Notably, several of them exhibit inferior performance compared to finetuning, suggesting that neglecting the complex inter-modal interactions during continual training negatively impacts the performance.
Our ConSurv method outperforms all other methods in the main metrics: average C-index and average C-index IPCW.
Furthermore, it achieves the highest BWT and FWT for C-index IPCW.
Other metrics of ConSurv are not the highest, since there is a trade-off between absolute performance and resistance to forgetting, thus they cannot comprehensively assess the effectiveness of ConSurv~\cite{Huang2023ConSlide}.
We list those metrics for reference, following previous works~\cite{Huang2023ConSlide, Lopez-Paz2017Gradient, Chaudhry2018Riemannian}.

\newcommand{\mySizeArchitectures}{3.6cm}

\begin{figure*}[!ht]
\centering
\subfloat[\small{BLCA}]{
            \includegraphics[height=\mySizeArchitectures]{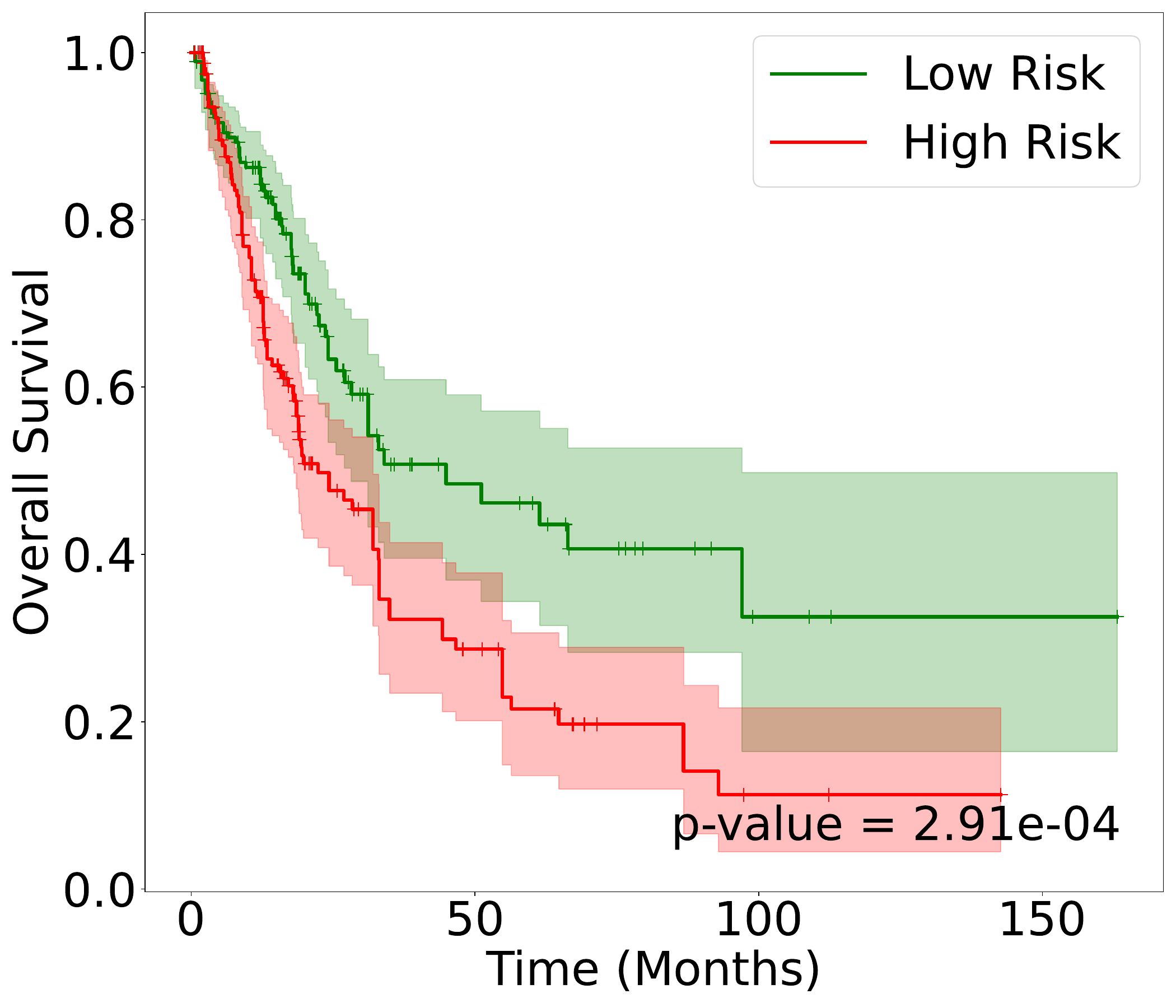} 
            \label{fig: a KM_analysis}
}
\subfloat[\small{UCEC}]{
            \includegraphics[height=\mySizeArchitectures]{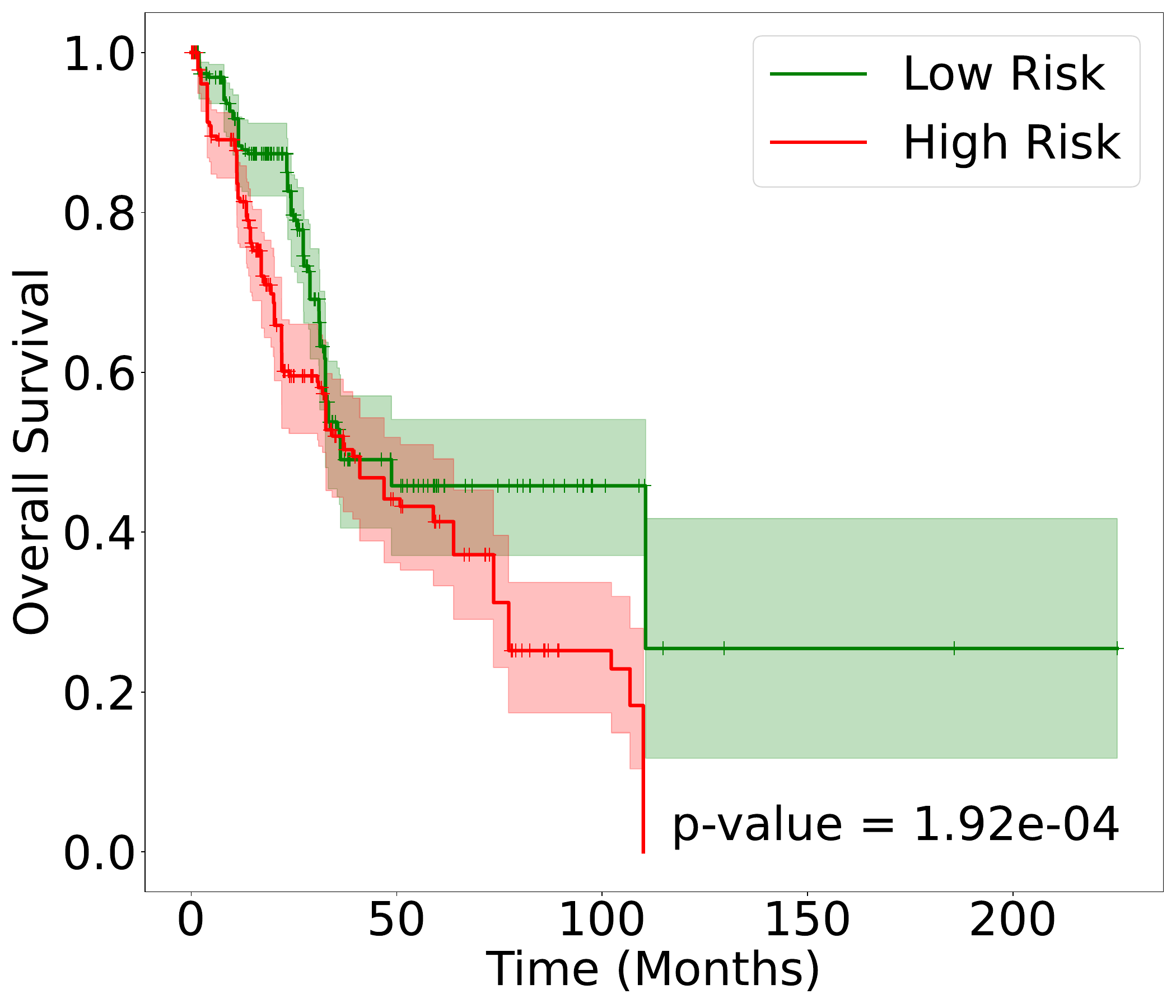}  
}
\subfloat[\small{LUAD}]{
            \includegraphics[height=\mySizeArchitectures]{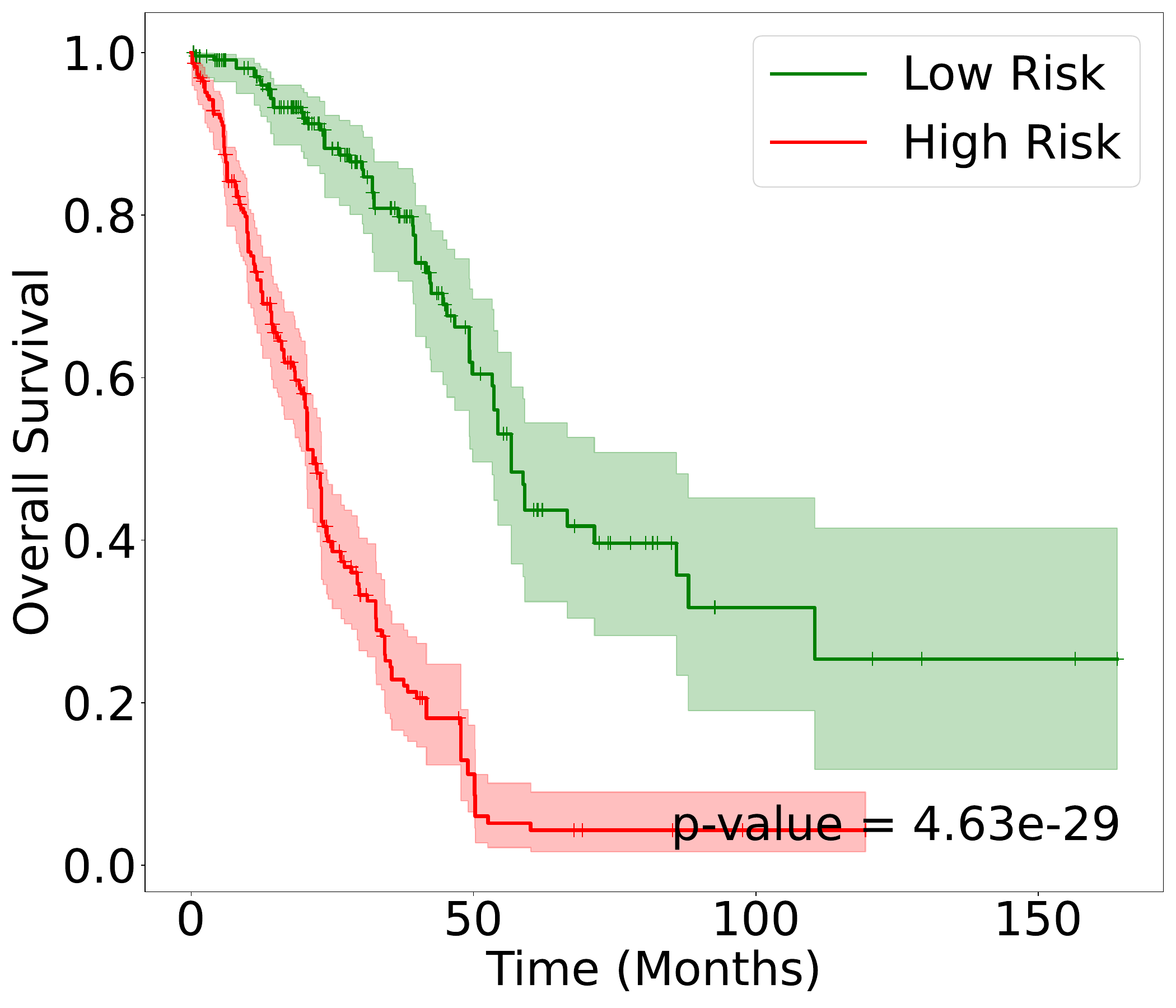}  
}
\subfloat[\small{BRCA}]{
            \includegraphics[height=\mySizeArchitectures]{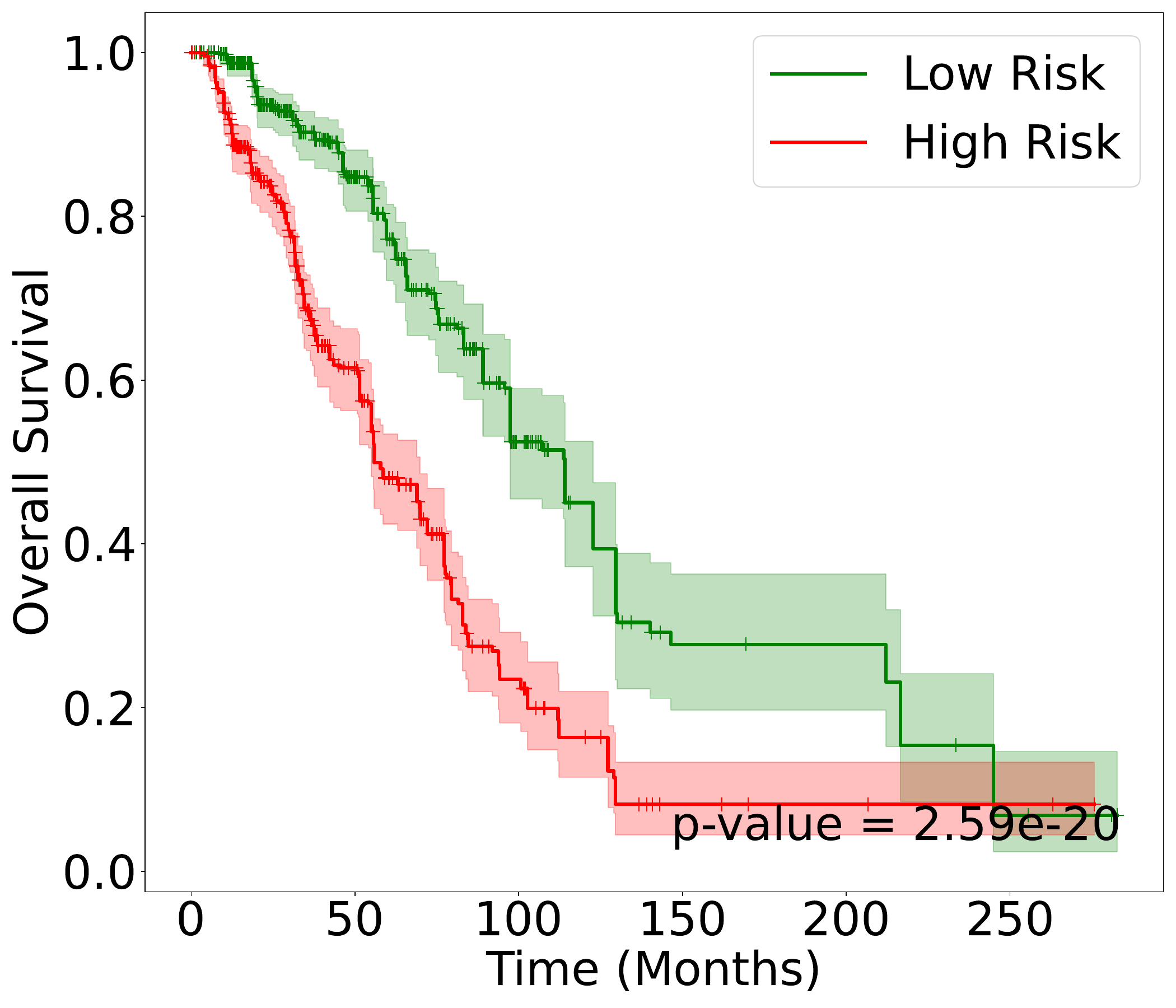}  
}
\caption{Kaplan-Meier curves of our ConSurv on Cancer4.}
\label{fig: KM_analysis}
\end{figure*}

\begin{figure}[!t]
    \centering
    \includegraphics[width=0.95\columnwidth,page=1]{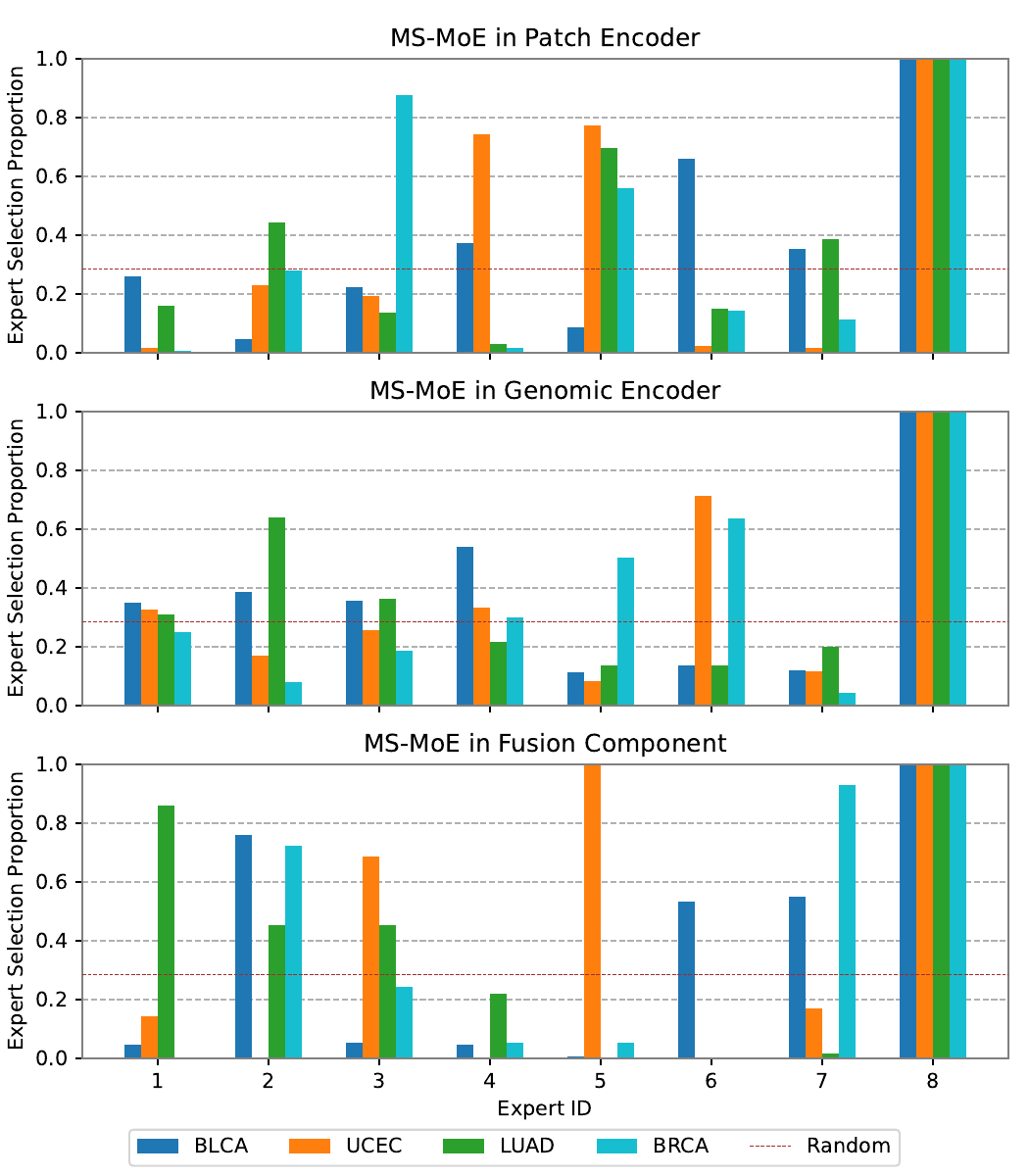}
    \caption{Proportion of each expert within the MS-MoE modules selected on inputs from different datasets. The brown dashed line represents the expected selection proportion under random sampling, which is 2/7. The last expert $\mathcal{E}_8$ functions as a shared expert and is always selected.}
    \label{fig: ms_moe_analysis}
\end{figure}

\subsection{Kaplan–Meier Analysis (RQ3.1)}
To further validate the differentiability of our ConSurv on each dataset in Cancer4, we perform a Kaplan–Meier analysis with the final model trained under the MMCL setting.
Based on the mean risk value of a dataset, we partition patients into low-risk and high-risk groups~\cite{Xiong2024Enhancing}.
The survival outcomes for all patients are visualized in~\Cref{fig: KM_analysis}. 
To assess the statistical significance of the difference between the two risk groups, we conduct a log-rank test, following~\cite{Xiong2024Enhancing}, with a p-value less than 0.05 considered statistically significant by convention.
As illustrated in~\Cref{fig: KM_analysis}, ConSurv successfully stratifies patients into low-risk and high-risk groups with high statistical significance, thus demonstrating its ability to learn and retain knowledge from multimodal data throughout the CL process while effectively mitigating catastrophic forgetting.

\subsection{Ablation Study (RQ3.2)}

We conduct an ablation study on our proposed FCR and MS-MoE modules to investigate their individual effects. The results are presented in~\Cref{tab: Ablation table}.
\subsubsection{The effects of MS-MoE.}
As shown in~\Cref{tab: Ablation table}, employing MS-MoE improves the average C-index and average C-index IPCW, compared to the finetuning baseline (the first row). Note that MS-MoE operates without the need for data replay. Thus, the buffer is not used.
This observation suggests that MS-MoE effectively facilitates learning of both shared and task-specific knowledge across datasets, while alleviating forgetting.

\subsubsection{The effects of FCR.}
The results in~\Cref{tab: Ablation table} indicate that utilizing FCR in isolation increases both the average C-index and the average C-index with IPCW, demonstrating its effectiveness. 
Furthermore, based on ConSurv with two modules, we investigate the performance when only the final fusion representation is constrained within FCR (the ``$\checkmark$(f)'' row in ~\Cref{tab: Ablation table}).
We discover that it already achieves the highest average C-index, compared with other CL methods, highlighting the efficacy of feature constraints.
Constraining features at the patch, genomics, and fusion levels collectively results in further improvements across most evaluation metrics (comparing the last two rows).
These findings suggest that the two modules collaborate effectively, ultimately leading to the best overall performance.

\subsection{MS-MoE Routing Analysis (RQ3.3)}
\label{sec: Routing Analysis}
This section presents an analysis of MS-MoE expert selection performed by the routers $\mathcal{R}_k$ for each dataset $\mathcal{D}_k$.
Our investigation aims to determine whether experts can acquire task-specific and shared knowledge across subsets of the datasets. 
As illustrated in~\Cref{fig: ms_moe_analysis}, we quantify the selection proportion for each expert on the validation datasets. 
The results reveal a diversity in expert preferences across different datasets.
Some experts specialize in learning knowledge important to a single dataset; for example, expert $\mathcal{E}_3$ within the patch encoder's MS-MoE focuses on BRCA.
Conversely, some experts are selected across multiple tasks, suggesting the acquisition of shared knowledge; for instance, expert $\mathcal{E}_6$ within the genomic encoder's MS-MoE has learned knowledge relevant to UCEC and BRCA.
This demonstrates MS-MoE's capacity for appropriate expert selection, which facilitates ConSurv's learning of multimodal knowledge throughout the CL process, providing further evidence for the effectiveness of MS-MoE.

\section{Conclusion}
In this work, we first explore the necessity of CL in multimodal survival prediction and quantify severe catastrophic forgetting in this new setting. 
We propose \textbf{ConSurv}, the \textbf{first} MMCL method for survival analysis, to tackle the challenges of forgetting and complex inter-modal interactions between gigapixel WSIs and genomics in different cancers. 
The proposed MS-MoE effectively learn shared and task-specific knowledge at different learning stages of the network, including WSI and genomic encoders and the modality fusion component. 
We design FCR to enhance learned knowledge by limiting feature deviation at multiple levels, including encoder-level features of two modalities and the fusion-level representations.
In addition, we establish the new MSAIL benchmark by integrating TCGA datasets and utilize it for evaluation.
Extensive experiments demonstrate that ConSurv surpasses other methods across multiple metrics, with a better trade-off between acquiring new knowledge and retaining previously learned information.
A detailed analysis of computational costs, limitations, and future work is provided in Appendix D.

\section*{Acknowledgments}
The work described in this paper was partially supported by the Research Grants Council of the Hong Kong Special Administrative Region, China (CUHK 2300246, RGC C1043-24G). 
We sincerely thank Professor Irwin King, for his unwavering support and expert guidance throughout every stage of this work.
We sincerely thank Professor Joseph~J.~Y.~Sung, for his invaluable and insightful suggestions that enhanced this paper, especially regarding the contrast between static and dynamic models, and the clinical importance of a dynamic model in survival prediction.

\bibliography{MyLibrary}

\clearpage %

\appendix

\section*{\centering Appendix}

\section{Related Work}

\subsection{Survival Prediction}
Deep learning techniques are employed for survival prediction tasks, with a notable impact on computational histopathology~\cite{Xiong2025Survey}.
Unimodal approaches predominantly fall into two categories: \textit{genomic-based} and \textit{WSI-based} approaches~\cite{Xiong2024Enhancing}.
Genomic-based strategies often leverage neural networks, including SNNs~\cite{Klambauer2017Selfnormalizing} and Transformers~\cite{Vaswani2017Attention}, for extracting relevant genomic features.
WSI-based methods typically involve segmenting gigapixel WSIs into smaller patches, framing the task as a Multiple Instance Learning (MIL) problem. Two main subcategories are \textit{instance-level} and \textit{embedding-level}~\cite{Xiong2024Enhancing}.
Instance-level algorithms~\cite{Campanella2019Clinicalgrade, Kanavati2020Weaklysupervised, Xu2019Camel} select instances for aggregation based on the prediction of each instance.
Embedding-level algorithms~\cite{Ilse2018Attentionbased, Lu2021Dataefficient, Shao2021Transmil, Zhang2022Dtfdmil, XiongDiagnose} first obtain embeddings for each patch and then aggregate them into a final representation for prediction.

Recently, research has increasingly focused on leveraging multimodal data, such as combining WSIs and genomics, since they can provide more comprehensive information about patients and diseases.
\textit{Tensor-based} methods utilize pre-defined tensor operations to fuse multimodal data, such as concatenation~\cite{Mobadersany2018Predicting} and bilinear pooling~\cite{Li2022HFBSurv}.
Subsequent works, such as MCAT~\cite{Chen2021Multimodal}, utilize co-attention mechanisms to learn genomic-guided representations for WSIs and maintain interpretability.
MOTCat~\cite{Xu2023Multimodalb} improves model performance and efficiency by using micro-batches of WSI patches, with optimal transport-based co-attention.
Other recent studies~\cite{Xiong2024MoME, Song2024Multimodal, Zhang2024Prototypical, Jaume2024Modeling} also utilize \textit{attention-based} mechanisms, aiming for more effective modeling of multimodal interactions to improve prediction accuracy.
However, all existing works focus on training on single datasets, while our work proposes an MMCL method to perform survival prediction for multiple cancer types.

\subsection{Continual Learning}
Continual learning (CL) endeavors to enable models to sequentially acquire knowledge from new data while preserving previously learned information, thus mitigating catastrophic forgetting. 
CL has three main scenarios: class-incremental, task-incremental, and domain-incremental learning~\cite{vandeVen2022Three}.
Unimodal CL methods generally fall into categories of regularization-based~\cite{Kirkpatrick2017Overcoming, Li2017Learning},
architecture-based~\cite{RusuRDSKKPH16,Yoon2020Scalable},
and replay-based methods~\cite{Chaudhry2019Tiny,Buzzega2020Dark,Zhang2024Influential}.

Recently, with the rapid growth of multimodal data~\cite{Wu2023Multimodal, Cui2024Recent}, the scope of CL has naturally expanded from unimodal to multimodal, similar to the trends observed in survival prediction.
However, directly adapting unimodal CL strategies to multimodal continual learning (MMCL) is not always feasible due to algorithm designs tailored for unimodal data. For instance, the image patch reorganization strategy of Conslide~\cite{Huang2023ConSlide} is incompatible with MMCL scenarios involving other modalities.
There have been numerous MMCL works designed for this setting, predominantly targeting vision-language tasks
These methods are often categorized as: 
regularization-based~\cite{He2023Continual,Zheng2023Preventing},
architecture-based~\cite{DelChiaro2020RATT,Yu2024Boosting},
replay-based~\cite{Zhang2023VQACL,Yan2022Generative}, 
and more recently, prompt-based approaches~\cite{DAlessandro2023Multimodal,Qian2023Decouple}.
Despite this progress, CL research focusing on modalities beyond vision and language remains limited~\cite{Yu2024Recent}. 
Specifically, there is a notable absence of CL studies on WSIs and genomics. Our work aims to present a dedicated study in this area and fill this gap with an MMCL method for survival analysis of multiple types of cancers.

\setlength{\tabcolsep}{5pt}
\begin{table*}[!t]
    \centering
\small
\begin{NiceTabular}{l|c|ccc|c|ccc}
    \CodeBefore
    \rowcolors{5}{gray!20}{}[cols=1-9,restart]
    \Body
    \toprule
    \multicolumn{1}{c|}{\multirow{2.5}{*}{Method}} & \multicolumn{4}{c|}{C-index}                                                               & \multicolumn{4}{c}{C-index IPCW}                      \\
    \cline{2-9}\addlinespace[0.5ex]
     & \textbf{Average} ($\uparrow$)            & Forget ($\downarrow$)         & BWT ($\uparrow$)                & FWT ($\uparrow$)                   & \textbf{Average} ($\uparrow$)              & Forget ($\downarrow$)            & BWT ($\uparrow$)                   & FWT ($\uparrow$)                  \\
  \midrule        
LwF                                        & 0.573±0.054          & 0.108±0.072          & -0.104±0.075          & \textbf{0.046±0.045} & 0.550±0.055          & 0.095±0.086          & \textbf{-0.010±0.140} & \textbf{0.107±0.112}           \\
    DER++                                       & 0.579±0.044          & 0.110±0.044          & -0.109±0.044          & 0.018±0.030          & 0.543±0.041          & 0.110±0.098          & -0.078±0.106          & 0.047±0.138      \\
    \textbf{ConSurv}                              & \textbf{0.597±0.026} & \textbf{0.096±0.026} & \textbf{-0.096±0.026} & -0.025±0.025         & \textbf{0.580±0.043} & \textbf{0.046±0.032} & -0.045±0.032          & -0.011±0.081     \\
\bottomrule
                                                
    \end{NiceTabular}
    \caption{Comparison results among different CL methods with Order-II.}
    \label{tab: new order}
    \end{table*}

\section{Preliminaries: Survival Prediction with WSIs and Genomics} 
\label{sec: Survival Prediction with WSIs and Genomics}
In this section, we present the problem formulation of survival prediction with the data modalities of WSIs and genomics.

\textbf{WSI.} 
Due to the prohibitive size of gigapixel WSIs, MIL is widely adopted to avoid using the whole WSIs as inputs.
In this approach, a WSI $\mathcal{I}$ is cropped into $n_P$ patches $\mathcal{P}$, where \mbox{$\mathcal{I}=[\mathcal{P}_1, \mathcal{P}_2, \dots, \mathcal{P}_{n_P}]$}.
Each patch $\mathcal{P}_i$ is then processed by a pre-trained vision model $f$ to extract features $\mathbf{P_i}=f(\mathcal{P}_i)$~\cite{Ilse2018Attentionbased,XiongDiagnose}.  
The resulting representation is {$\mathbf{P}=[\mathbf{P}_1, \mathbf{P}_2, \dots, \mathbf{P}_{n_P}] \in \mathbb{R}^{n_P \times d_P}$}, where $d_P$ is the feature dimension.

\textbf{Genomics.}
The genomic data are grouped into six functional groups that are closely relevant to cancer progression: Tumor Suppression, Oncogenesis, Protein Kinases, Cellular Differentiation, Transcription, and Cytokines and Growth~\cite{Chen2021Multimodal, Xu2023Multimodalb, Xiong2024MoME}.
The genomic data is $\mathbf{G}=[\mathbf{G}_1, \mathbf{G}_2, \dots, \mathbf{G}_{n_{G}}]$, where $n_{G}=6$ is the number of groups and $\mathbf{G}_i \in \mathbb{R}^{d_{G_i}}$ is the dimension of the $i$-th genomic group.

\textbf{Survival prediction.}
A special source of difficulty in survival prediction lies in the possibility that some outcomes of the events are not observed~\cite{Cox1984Analysis}. We use $c\in\{0,1\}$ to denote the censorship status, indicating whether the survival time results from the observed patient's death (uncensored) or the last known patient follow-up without observed outcomes (censored).
The survival time $t$ in continuous time scale are partitioned into non-overlapping $n_B$ bins with labels $B=\{ i \in \mathbb{N} \mid i < n_B \}$ and time intervals \mbox{$\{[t_i, t_{i+1}) | i\in B\}$}, where fixed boundaries are determined by the survival time of uncensored patients~\cite{Chen2021Multimodal}. 
For the $i$-th patient with original survival time $t^{(i)}$, let $Y^{(i)}\in B$ be the resulting bin, which is also the discrete time ground truth label, and $t_{Y^{(i)}}$ is thus the left endpoint of the bin.
The dataset after processing is:
\begin{equation}
    \mathcal{D} = \{(\mathbf{P}^{(i)}, \mathbf{G}^{(i)}, Y^{(i)}, c^{(i)}): i\in \mathbb{N}, 1\leq i\leq N\},
\end{equation} 
where $N$ is the number of samples in the dataset $\mathcal{D}$.
Let $T$ denote the time until death, and $(\mathbf{P}, \mathbf{G}, Y, c)$ represent a sample. A model parameterized by $\bm{\theta}$ outputs the probability associated with each bin.
It models the hazard function $h(t|\mathbf{P}, \mathbf{G})$, representing the probability of death for a patient right after the time point $t$, defined as~\cite{Chen2021Multimodal, Xiong2024MoME}:
\begin{equation}
    h(t|\mathbf{P}, \mathbf{G}) = P(T = t|T \geq t, \mathbf{P}, \mathbf{G}) \in [0, 1], 
\end{equation} 
where $P(\cdot)$ denotes probability.
The survival function, $S(t|\mathbf{P}, \mathbf{G})$, represents the probability of surviving beyond time $t$, and is derived from the hazard function. 
Define $t_{-1}=-\infty$ and $S(t_{-1}|\mathbf{P}, \mathbf{G})=1$, i.e. all patients are alive beyond time $t_{-1}$.
Using discrete time bins, we can define the survival function for $t_r$, where $r \in B$~\cite{Chen2021Multimodal}:
\begin{align}
    S(t_r| \mathbf{P}, \mathbf{G}) = P(T > t_r | \mathbf{P}, \mathbf{G}) = \prod_{u=0}^{r} (1 - h(t_u | \mathbf{P}, \mathbf{G})). 
\end{align}
Let the hazard function modeled by $\bm{\theta}$ be $h_{\bm{\theta}}$ and the corresponding survival function be $S_{\bm{\theta}}$.
We employ the Negative Log-Likelihood (NLL) loss and utilize both uncensored and censored data to train the model. $\ell_s$ on one sample is defined as~\cite{Zadeh2020Bias,Chen2021Multimodal}:
\begin{align}
    \ell_s(\mathbf{P}, \mathbf{G}, Y, c; \bm{\theta}) &= - (1 - c) (\log S_{\bm{\theta}}(t_{Y-1}|\mathbf{P}, \mathbf{G}) \\
    &+ \log h_{\bm{\theta}}(t_Y|\mathbf{P}, \mathbf{G})) \notag \\
    & - c (1 - \alpha_s) \log S_{\bm{\theta}}(t_{Y}|\mathbf{P}, \mathbf{G}),  
\end{align} 
where $\alpha_s$ denotes a weighting factor that balances uncensored and censored loss components.
The goal is to train a model, minimizing the loss $\mathcal{L}_s$ on the given dataset $\mathcal{D}$:
\begin{align}
    \mathcal{L}_s(\mathcal{D};\bm{\theta}) &= \mathbb{E}_{(\mathbf{P}, \mathbf{G}, Y, c) \sim \mathcal{D}} [\ell_s(\mathbf{P}, \mathbf{G}, Y, c; \bm{\theta})], \\
    \bm{\theta}^{*}&=\underset{\bm{\theta}}{\operatorname{argmin}} \ \mathcal{L}_s(\mathcal{D};\bm{\theta}).
\end{align}

\section{Experiments}

\subsection{Experimental Settings}
We train a model with Cancer4 merged as one dataset (Joint), which can be regarded as an upper-bound and does not participate in the comparison.
As a lower-bound reference, we perform direct sequential finetuning of a model on Cancer4 (Finetune).
We compare our proposed ConSurv against several state-of-the-art CL approaches: regularization-based methods LwF~\cite{Li2017Learning} and EWC~\cite{Kirkpatrick2017Overcoming}, an architecture-based method T-LoRA, which is our adaptation of task-specific LoRA~\cite{Hu2022LoRA} to this setting,
and replay-based methods ER~\cite{Chaudhry2019Tiny}, DER~\cite{Buzzega2020Dark}, DER++~\cite{Buzzega2020Dark}, MOSE~\cite{Yan2024Orchestrate}, MOE-MOSE~\cite{Yan2024Orchestrate}, and IMEX-Reg~\cite{Bhat2024IMEXReg}.
The replay buffer is the same for all replay-based methods (32 WSIs). 
We re-implement these CL methods under the multimodal setting for survival prediction.

\subsection{Evaluation Details}
All methods are evaluated using the metrics detailed in~\Cref{sec: Metrics} with five-fold cross-validation.
For each dataset, the model is trained for 20 epochs, and the checkpoint achieving the best validation C-index within these epochs is saved for the subsequent continual training on the next dataset.
The means and standard deviations of metrics for each method after training on all datasets are reported.

\subsection{Experiment Implementation Details}
WSIs are split into patches of $256\times256$ pixels at $20\times$ magnification.
We use ResNet-50~\cite{He2016Deep} pre-trained on ImageNet to extract features.
Following previous works~\cite{Chen2021Multimodal, Xu2023Multimodalb, Xiong2024MoME}, we utilize four bins to discretize the survival time ($n_B=4$), with the bin boundaries determined by the quartiles of the survival times of uncensored patients.
To address variations in the dimensions of genomic features across different datasets, we pre-process the data and use zero padding to ensure a consistent input dimension for the model. Adam~\cite{Kingma2014Adam} optimizer is employed in our experiments. 
The learning rate and weight decay are set to $2\times10^{-4}$ and $1\times10^{-5}$, respectively~\cite{Xu2023Multimodalb}. The micro-batch technique~\cite{Xu2023Multimodalb} is utilized, with a micro-batch size of 4,096~\cite{Xiong2024MoME}.
We follow the TIL setting, and task labels are available at inference.
For each new dataset, we use a different head to perform survival prediction.
For the hyperparameters, $\beta=0.5$ is adopted from the strong baseline DER++~\cite{Buzzega2020Dark} for a fair comparison.
The value of $\alpha=2.4 \times 10^{-3}$ was determined by analyzing and balancing the initial magnitudes of the loss terms, and we also have tried values in $[1e-4,1e-2]$.
In MS-MoE, the total number of experts is set to eight ($n_E=8$), following the successful configuration of the recent MoE-based model Mixtral~\cite{Jiang2024Mixtrala}. One of them is the shared expert.
Two experts among the rest are sparsely selected, following~\cite{Jiang2024Mixtrala}. 
All experiments are conducted with PyTorch on a NVIDIA A100 GPU.

\subsection{Changing Task Orders}
We conduct experiments to further validate the robustness of our ConSurv.
We utilize an alternative task order of cancer datasets, referred to as Order-II: LUAD, BLCA, UCEC, and BRCA.
We compare the performance of the top three methods (ConSurv, DER++, and LwF) and show results in~\Cref{tab: new order}.
We observe that our ConSurv achieves the highest performance in the main metrics: average C-index and average C-index IPCW.
Furthermore, ConSurv obtains the best (lowest) Forgetting score. 
The results demonstrate the robustness of our method.

\section{Discussion}
\subsection{Module Analysis}
\label{sec: Module Analysis}
We analyze the resource requirements of our proposed MS-MoE and FCR modules in the following aspects.

\textbf{Parameters.} The MS-MoE module introduces 8.5M additional parameters, while the FCR module adds no parameters to the model.

\textbf{Memory.} MS-MoE has memory of 32.5MB, accounting for just 14\% of the total model memory (232.13MB).
The replay buffer of FCR resides on the CPU, and data is moved to the GPU on demand, thus having a negligible impact on GPU memory.

\textbf{Computation.} For an input of dimension $D$ and output of dimension $L$, an MS-MoE module introduces an additional \mbox{$16D+6DL+6L^2$} FLOPs. 
The FCR module processes the same number of replayed samples as the strong baseline DER++~\cite{Buzzega2020Dark}. 
Assuming a constant sample processing time, they have the same time complexity with respect to samples. In practice, ConSurv (6.62h) is faster than IMEX-Reg (8.68h) and only marginally slower than DER++ (6.05h), while achieving superior performance.

\subsection{Limitations}
Our approach has several limitations. As discussed in Appendix~\ref{sec: Module Analysis}, both the FCR and MS-MoE modules introduce additional computational overhead compared to a simple finetuning baseline. Furthermore, as a replay-based method, FCR has potential privacy concerns associated with storing representations of past data. 

\subsection{Future Work}
To address the aforementioned limitations, future work could explore replay-free CL methods to mitigate privacy concerns. For practical deployment in clinical systems where patient privacy is important, integrating our CL framework with federated learning~\cite{Yang2024Federated, Zhang2025PFedMxF} presents a promising research direction.
Additionally, to address the high cost of medical image annotation, exploring active learning~\cite{Li2024Surveya, Song2023No} to efficiently leverage unlabeled data is another critical future direction.

\end{document}